\documentclass[10pt, a4paper]{article}

\usepackage[final]{lrec2026} 
\usepackage[nolist]{acronym}
\usepackage{enumitem}
\usepackage{tabularx}
\usepackage{booktabs}
\usepackage{array}
\usepackage{multirow}
\usepackage{amsmath}
\usepackage{longtable}
\usepackage{graphicx}
\usepackage{xcolor} 
\usepackage{inconsolata}
\usepackage{listings}
\usepackage[most]{tcolorbox}
\usepackage{float}

\newtcblisting{mycode}[2][text]{%
  listing only,
  colback=gray!15,
  colframe=black,
  coltitle=white,
  colbacktitle=black,
  fonttitle=\bfseries,
  title={#2},
  listing options={
    language=#1,
    basicstyle=\ttfamily\footnotesize\color{black},
    breaklines=true,
    breakautoindent=true,
    breakindent=0em,
    numbers=none,
    keywordstyle=\color{black},
    commentstyle=\color{black},
    stringstyle=\color{black},
    tabsize=2
  }
}

\title{Is this Idea Novel? An Automated Benchmark for Judgment of Research Ideas \raisebox{-0.15em}{\includegraphics[height=1.0em]{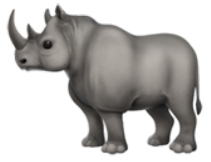}}}

\name{Tim Schopf$^{1,2}$ and Michael Färber$^1$}

\address{$^1$TU Dresden \& ScaDS.AI Dresden/Leipzig, Germany \\
         $^2$National Institute of Informatics, Tokyo, Japan \\
         \{tim.schopf, michael.färber\}@tu-dresden.de\\}

\abstract{Judging the novelty of research ideas is crucial for advancing science, enabling the identification of unexplored directions, and ensuring contributions meaningfully extend existing knowledge rather than reiterate minor variations. However, given the exponential growth of scientific literature, manually judging the novelty of research ideas through literature reviews is labor-intensive, subjective, and infeasible at scale. Therefore, recent efforts have proposed automated approaches for research idea novelty judgment. Yet, evaluation of these approaches remains largely inconsistent and is typically based on non-standardized human evaluations, hindering large-scale, comparable evaluations. To address this, we introduce \mbox{\texttt{RINoBench}}, the first comprehensive benchmark for large-scale evaluation of research idea novelty judgments. It comprises 1,381 research ideas derived from and judged by human experts as well as nine automated evaluation metrics designed to assess both rubric-based novelty scores and textual justifications of novelty judgments. Using this benchmark, we evaluate several state-of-the-art large language models (LLMs) on their ability to judge the novelty of research ideas. Our findings reveal that while LLM-generated reasoning closely mirrors human rationales, this alignment does not reliably translate into accurate novelty judgments, which diverge significantly from human gold standard judgments—even among leading reasoning-capable models. Data and code available at: \url{https://github.com/TimSchopf/RINoBench}.
 \\ \newline }

\begin{acronym}
    \acro{llm}[LLM]{large language model}
    \acro{mae}[MAE]{Mean Absolute Error}
    \acro{rinobench-emoji}[\mbox{\texttt{RINoBench\raisebox{-0.15em}{\includegraphics[height=1em]{figures/rhino_apple.pdf}}}}]{\textbf{R}esearch \textbf{I}dea \textbf{No}velty Judgment \textbf{Bench}mark}
    \acro{rino}[\mbox{\texttt{RINoBench}}]{\textbf{R}esearch \textbf{I}dea \textbf{No}velty Judgment \textbf{Bench}mark}
\end{acronym}

\begin{document}

\maketitleabstract

\section{Introduction}
Judging the novelty of research ideas is fundamental to fostering scientific discovery and ensuring that new works meaningfully advance a field rather than reproducing existing results with minor variations that contribute little new insight. Hence, effective novelty judgment helps researchers identify unexplored directions, develop original contributions, and ultimately drive scientific progress. However, manually judging the novelty of a research idea requires a comprehensive review of previous work to determine whether the same or similar ideas have already been explored. With the rapid growth of scientific literature \cite{doi:10.1126/science.aao0185}, this manual process has become increasingly challenging for researchers in terms of both time and cognitive effort. Moreover, novelty judgments are inherently subjective. Experts can often identify when two ideas are similar \cite{picard25} but struggle to articulate what makes an idea truly novel \cite{shahid-etal-2025-literature}. In addition, such judgments are influenced by an individual’s prior knowledge, intuition, and familiarity with the relevant literature \cite{10.1115/1.4041856,picard25}. To address these challenges, automated approaches have been proposed to support and enhance research idea novelty judgments.

\begin{figure}[t]
    \centering
    \includegraphics[width=1\columnwidth]{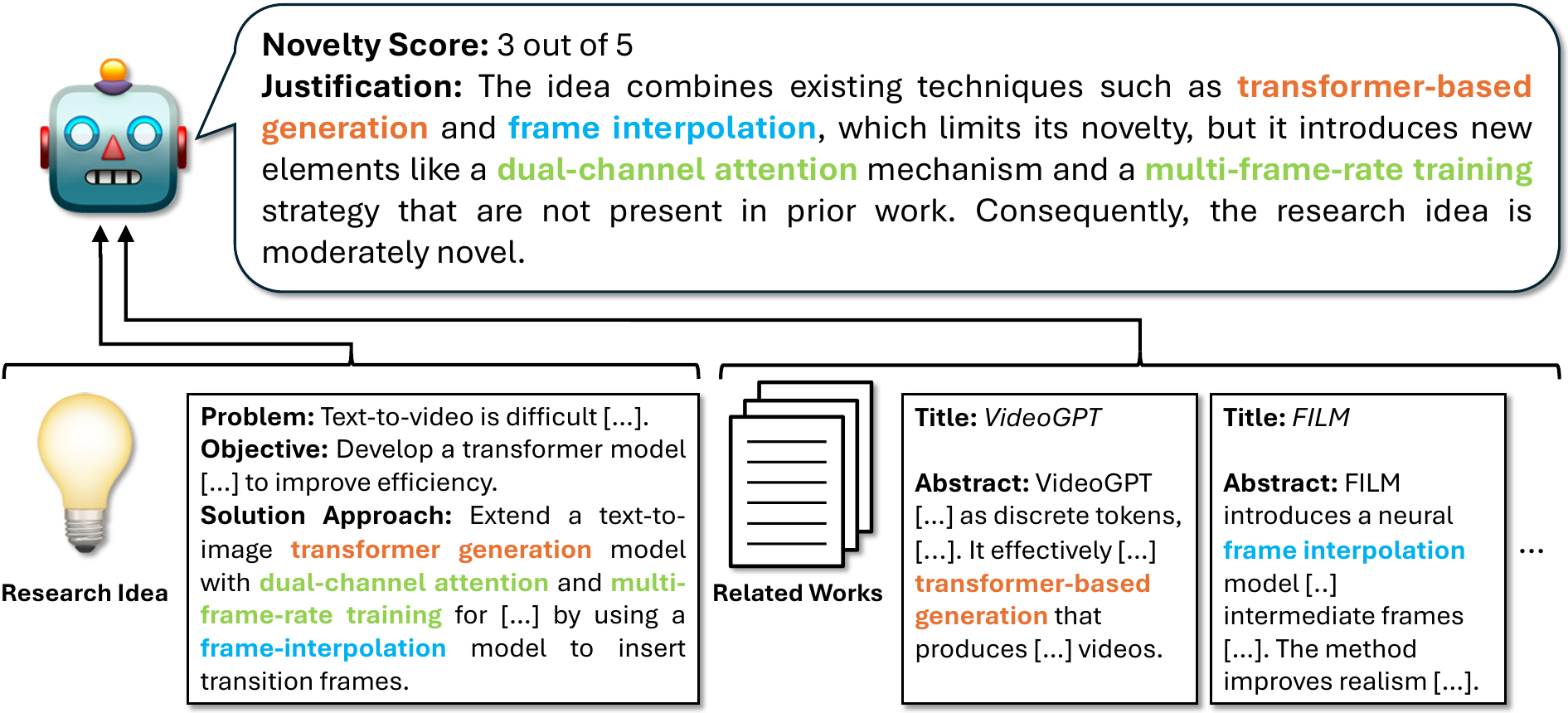}
    \caption{The task setup of \acs{rino}. Given a research idea and its related works, a model must judge the novelty of the idea according to a five-point rubric. In addition, the model must provide a textual justification for its judgment, grounded in a comparison between the proposed research idea and the related works.}
    \label{fig:task_example}
\end{figure}

Recent work has used \acp{llm} to automatically judge the novelty of research ideas \cite{lu2024aiscientistfullyautomated,si2025can,li-etal-2025-chain-ideas,su-etal-2025-many,gottweis2025aicoscientist}. However, these approaches do not ground their rationales in prior literature and struggle with subtle linguistic variation, leading to the misclassification of well-established ideas as novel \cite{10.1145/3769733.3769747,gupta-pruthi-2025-glitters,wang2025thetreetracinghistoricalevolution}. Furthermore, many \ac{llm}-based approaches restrict their outputs to binary classifications (novel vs. not novel) \cite{lu2024aiscientistfullyautomated,li-etal-2025-chain-ideas,shahid-etal-2025-literature,su-etal-2025-many}, overlooking the nuanced and gradual nature of novelty judgments. Moreover, most automated approaches provide only final predictions without offering interpretable explanations or justifications supporting their decisions. This lack of transparency reduces their practical utility, as researchers cannot review or learn from opaque judgments, hindering their ability to refine research ideas toward greater novelty. Finally, the fundamental limitations and differences between existing automated research idea judgment approaches make meaningful comparisons difficult. This problem is exacerbated by the fact that current evaluations of automated research idea judgment approaches are mainly based on non-standardized manual evaluations \cite[\textit{inter alia}]{si2025can,gottweis2025aicoscientist}, hindering large-scale, systematic comparisons. 


To address these limitations, we introduce the \ac{rinobench-emoji}, the first \textit{comprehensive} and \textit{reproducible} benchmark for the automatic evaluation of research idea novelty judgments. Using this benchmark, we conduct the first large-scale benchmarking study of current state-of-the-art \acp{llm}, evaluating their ability to judge the novelty of research ideas. We reveal that \textit{while \acp{llm} often generate reasoning patterns similar to human experts, they fail to consistently translate these rationales into accurate novelty judgments}.

Our main contributions are: 
\begin{itemize}[topsep=0pt, partopsep=0pt, parsep=0pt, itemsep=3pt]
    \item \acs{rino}, a comprehensive benchmark for systematically evaluating research idea novelty judgments, comprising \textbf{1,381 research ideas derived from and judged by human experts} as well as \textbf{nine automated evaluation metrics} designed to assess both rubric-based novelty scores and textual justifications of novelty judgments.
    \item A study investigating several state-of-the-art \acp{llm} on their ability to judge the novelty of research ideas, involving a systematic analysis of the strengths and limitations of current state-of-the-art \acp{llm} performing this task.
\end{itemize}

\section{Related Work}

Automated methods for judging novelty in scientific literature have advanced significantly in recent years. Early approaches measured novelty via atypical combinations of cited references \cite{doi:10.1126/science.1240474}, constructed historical co-occurrence matrices and derived journal vectors, where lower cosine similarity indicates greater novelty \cite{WANG20171416}, and relied on lexical similarity \cite{wang2019idea,SARICA2020112995}. However, such approaches are inherently limited in their ability to capture paraphrased, conceptually equivalent, or closely related ideas, as they reduce novelty to patterns of statistical co-occurrence or lexical overlap rather than accounting for their semantic relationships. Subsequent work using semantic embeddings \cite{gómezpérez2022artificialintelligencenaturallanguage} enhanced the ability to identify deeper conceptual relations, but remains constrained to surface-level semantic comparisons \cite{mysore-etal-2022-multi}. More recently, \citet{https://doi.org/10.1002/asi.70005} combined human and \ac{llm} knowledge for novelty evaluation. In addition, retrieval-augmented \ac{llm} approaches have emerged as promising alternatives \cite[\textit{inter alia}]{yang-etal-2024-large-language,bougie2024generativeadversarialreviewsllms,lu2024aiscientistfullyautomated,radensky2024scideator,si2025can,liu2025harnessinglargelanguagemodels,su-etal-2025-many,wang2025scipipllmbasedscientificpaper,baek-etal-2025-researchagent,zhang2025noveltybench,tang2025airesearcher,li-etal-2025-chain-ideas}. However, these approaches typically treat novelty judgment of research ideas as an intermediate step within a broader AI-assisted scientific discovery pipeline. As a result, and further exacerbated by the lack of automated benchmarks, these works either omit a dedicated and systematic evaluation of their novelty judgments or rely on costly and often small-scale human evaluations. Complementary to this, \citet{wen2025predicting} introduce a large-scale benchmark dataset designed to predict which of two research ideas performs better on a given set of benchmarks, but they do not address the task of novelty judgment. In addition, \citet{WU2025126778} examine which sections of research papers are most informative for novelty judgment. The work most closely related to ours provides the only publicly available evaluation dataset to date for judging the novelty of research ideas \cite{shahid-etal-2025-literature}. However, this dataset is limited to 51 manually annotated research ideas with binary labels (novel vs. not novel) and does not include any evaluation of textual novelty judgment justifications.


\section{On the Notion of Novelty}

\textit{Novelty} is a fundamental concept in scientific research, which has been extensively characterized in existing literature. \citet{ARTS2021104144} consider novelty as the uniqueness of specific knowledge elements, whereby the introduction of previously unknown elements indicates novel information. Further, \citet{Foster2021} define novelty as the extent to which a proposed contribution diverges from the existing scientific literature. Importantly, novelty is not limited to entirely new knowledge. An idea can also be considered novel if it represents a previously unobserved combination of known knowledge elements or applies them to new contexts \cite{doi:10.1287/mnsc.2015.2285,shahid-etal-2025-literature}.

Closely related to novelty is the concept of \textit{originality}. It refers to the generation of new ideas, methods, conclusions, or other valuable outputs that deviate from existing paradigms and can inspire further innovation \cite{Shibayama2019,HOU2022101306}. In practice, however, distinguishing originality from novelty is challenging \cite{doi:10.1177/000312240406900203}, leading to the frequent interchangeable use of these terms \cite{WU2025126778}.

In essence, novelty, often used interchangeably with originality, is a fundamental driver of scientific progress, providing the foundation for both innovation and disruptive advances. While a novel idea frequently entails introducing a previously unseen element of knowledge, it can also emerge from a previously unexplored combination of existing knowledge in innovative ways.

\section{Benchmark}

\acs{rino} unifies approaches for judging the novelty of research ideas by formalizing the task, illustrated in Figure \ref{fig:task_example}, as the process of comparing a proposed research idea with existing work to identify meaningful differences. Further, the task requires predicting a rubric-based novelty score (1–5) alongside a textual justification that grounds the judgment in related literature. 

\subsection{Data}

Collecting a comprehensive dataset through dedicated workshops or user studies is prohibitively expensive and practically infeasible due to the complexity of the task. Human experts would need to generate novel research ideas, and other experts would then need to evaluate them. Both tasks impose a high cognitive load and require substantial domain expertise, meaning that each instance demands significant time and effort. Moreover, the pool of qualified participants is small, making recruitment difficult. Consequently, prior data collection efforts of this kind have been limited in scale, typically yielding only around 50 human-generated research ideas \cite{si2025can,shahid-etal-2025-literature}.

To overcome these limitations, we adopt a different strategy by leveraging publicly available data from OpenReview. Specifically, peer reviews from ICLR 2022 and ICLR 2023 provide a rich source of information: human experts have already submitted papers based on their research ideas, which have been explicitly evaluated by other human experts using rubric-based novelty scores and corresponding textual justifications. By processing and enriching this peer review data, we construct a high-quality dataset for studying research idea novelty judgment.

\subsubsection{Data Collection \& Processing}
We collect all publicly available ICLR 2022 and ICLR 2023 submissions and their corresponding reviews from OpenReview, yielding 6,410 papers with associated reviewer feedback. Each submission was evaluated by approximately three expert reviewers, who rated the novelty of the research using a rubric-based numerical scale and provided brief textual justifications. Specifically, reviewers assessed two novelty dimensions: ``\textit{Technical Novelty and Significance}'' and ``\textit{Empirical Novelty and Significance}''. We use both dimensions in our dataset. Since human novelty judgments are inherently subjective and may vary substantially, we filter out all submissions where the maximum disagreement between reviewers exceeds one point within and across both novelty dimensions. This results in a filtered dataset of 3,535 paper submissions with high inter-reviewer agreement. 

To obtain a single gold-standard novelty score for each paper, we average all reviewer scores across both novelty dimensions. The resulting decimal values, however, are difficult to interpret and predict. To address this, we transform them into whole numbers on a unified 1–5 scoring rubric by binning the averaged values into five intervals. This 1-5 scoring rubric, as shown in Table \ref{tab:novelty-rubric}, offers an intuitive and standardized measure of novelty, featuring a clear midpoint, balanced polarity, and nuanced gradation, consistent with conventions commonly used in user studies. 

\begin{table}[h!]
\small
\centering
\renewcommand{\arraystretch}{1.1} 
\begin{tabularx}{\columnwidth}{|c|X|}
\hline
\multicolumn{1}{|c|}{\textbf{Score}} & \multicolumn{1}{c|}{\textbf{Degree of Novelty}} \\
\hline
\multirow{1}{*}{1} & The idea is not novel. All aspects already exist in prior work. \\ \hline
\multirow{2}{*}{2} & The idea is marginally novel. It represents only a minor variation of existing work. \\ \hline
\multirow{3}{*}{3} & The idea is somewhat novel. Aspects already exist in prior work. However, it might combine known approaches in new ways, apply them to new contexts, or propose incremental updates. \\ \hline
\multirow{2}{*}{4} & The idea is novel. It introduces new aspects not present in existing work. \\ \hline
\multirow{3}{*}{5} & The idea is highly innovative and novel. It is not present in existing work and potentially encourages new thinking or opens up new research directions. \\ \hline
\end{tabularx}
\caption{Novelty Judgment Rubric}
\label{tab:novelty-rubric}
\end{table}

Our next data processing step focuses on transforming submitted papers into concise research ideas by systematically identifying and reformulating their core ideas and contributions. Specifically, we provide an \ac{llm} with paper titles, abstracts, and reviewer summaries as context to distill the most salient information and produce structured and concise representations of the underlying research ideas. For this and all subsequent LLM-based data processing steps, we use the \textit{GPT-OSS-120B} \cite{openai2025gptoss120bgptoss20bmodel} model. In this step, the model is prompted to analyze the provided context, identify the key elements that define a paper’s research idea, and output a structured JSON representation capturing the core facets of the research idea. A central challenge in this process involves obtaining a reproducible and comparable representation of research ideas that contains all the information necessary for novelty understanding and judging. To this end, we adapt existing research idea templates \cite{si2025can,shahid-etal-2025-literature} to structured presentations consisting of the following aspects:

\begin{itemize}[topsep=0pt, partopsep=0pt, parsep=0pt, itemsep=3pt]
    \item \textbf{Problem statement:} A detailed description of the core research problem(s) or question(s) addressed.
    \item \textbf{Objective:} A clear articulation of the research aim(s) or intended outcomes.
    \item \textbf{Solution approach:} A detailed description of the proposed methods or approaches designed to solve the problem and achieve the stated objectives.
\end{itemize}

Following this, we synthesize the individual reviewer justifications for their novelty scores into a single, coherent justification aligned with each gold-standard novelty judgment score. To achieve this, we provide an \ac{llm} with the reviewers’ textual comments, the generated research idea, the assigned gold novelty score, and the novelty judgment rubric as context. The model is then prompted to identify the reviewers’ arguments justifying their given novelty scores for a research idea and to integrate them into a unified, coherent justification that explains the rationale behind the gold novelty score.

Finally, after involving an \ac{llm} in earlier stages of data processing and accounting for their tendency to produce hallucinations and inaccuracies, our final step focuses on two objectives: enriching research ideas with relevant related works to enable grounded novelty judgments, and enforcing strict quality control to ensure that only high-quality samples are included in the final dataset. To this end, we first obtain related works by retrieving the titles and abstracts of publications cited in the introduction and related work sections of paper submissions. We extract the relevant paper sections from the PDF submissions using Nougat \cite{blecher2024nougat} and obtain the works cited therein via Semantic Scholar \cite{kinney2025semanticscholaropendata}. Citations in other sections are ignored, as they typically include references to evaluation metrics or datasets that are irrelevant for novelty assessment. Next, we apply a quality filtering step. Research ideas are excluded if fewer than five related works are retrieved, typically due to missing indexing in Semantic Scholar. We then use an \ac{llm} to verify the formal correctness of the research idea, ensuring it is written in the first person, not as a summary of multiple reviews, and contains no explicit numerical novelty scores. Additionally, we assess whether all arguments in the synthesized novelty justifications are fully grounded in the corresponding research ideas and related works. Ungrounded justifications may arise not only from \ac{llm}-induced hallucinations during the synthesis of reviewer arguments, but also when reviewers use arguments derived from related works that are not cited in the corresponding submitted paper, or when related works are not available via Semantic Scholar. To assess the grounding of these justifications, we use an \ac{llm} to verify that every argument in the textual novelty justification is grounded in either the research idea or in the retrieved related works. In particular, the \ac{llm} extracts all arguments from the novelty justification pertaining to the novelty aspects of the research idea and verifies whether each novelty argument is grounded in the idea. In addition, the \ac{llm} extracts all arguments involving comparisons to related work and ensures that each argument is grounded in at least one retrieved title and abstract. Only samples with a formally correct research idea and a fully grounded novelty justification are included in the final data set.

\subsubsection{Dataset}

Our final dataset consists of 1,381 research ideas, each paired with rubric-based novelty scores, corresponding textual novelty judgment justifications, and an average of 25.23 titles and abstracts of related works. We perform a stratified 80:20 train-test split on the dataset, yielding the data distribution presented in Table \ref{tab:dataset}.

\begin{table}[ht!]
    \footnotesize
    \centering
    \renewcommand{\arraystretch}{1.0} 
    \begin{tabular}{|c||c|c|c|}
    \hline
    \textbf{Novelty Score} & \textbf{\#training} & \textbf{\#test} & $\sum$ \\ 
    \hline\hline
    \textbf{1} & 60 & 15 & 75 \\ 
    \hline
    \textbf{2} & 239 & 60 & 299 \\  
    \hline
    \textbf{3} & 349 & 87 & 436 \\ 
    \hline
    \textbf{4} & 322 & 81 & 403 \\ 
    \hline
    \textbf{5} & 134 & 34 & 168 \\ 
    \hline\hline
    $\sum$ & 1,104 & 277 & 1,381 \\ 
    \hline
    \end{tabular}
    \caption{Class distributions within \acs{rino}.}
\label{tab:dataset}
\end{table}

\subsection{Evaluation Metrics}

This section outlines the metrics used in \acs{rino} to evaluate both the predicted novelty scores and the generated textual justifications for the novelty judgments.

\subsubsection{Novelty Score Metrics} \label{sec:novelty_score_metrics}

\paragraph{$\mathbf{F}_{\mathbf{1}}$} We treat novelty score prediction as a classification task and use ${F}_{1}$ as the primary evaluation metric. Specifically, we employ macro-averaged ${F}_{1}$ to ensure that each degree of novelty is weighted equally. This approach disregards the actual category imbalance and treats all categories as equally important, consistent with how they are valued in practice. Additionally, we report the ${F}_{1}$ scores for each novelty category individually to evaluate how a model performs across different categories, identifying where it performs particularly well or poorly.

\paragraph{\ac{mae}} Since novelty scores in this task are not limited to discrete categories but also represent rubric-based values, \ac{mae} allows us to evaluate the magnitude of deviation between predicted and gold scores. This provides insight not only into whether correct novelty scores are predicted, but also into how far the predictions diverge from the gold standard. By evaluating the average distance of predictions from the gold scores, we can determine if a model's predictions are reasonably close or significantly misaligned with the expected outcomes.

\subsubsection{Justification Metrics} \label{sec:justification_metrics}

\begin{figure}[h!]
    \centering
    \includegraphics[width=1\columnwidth]{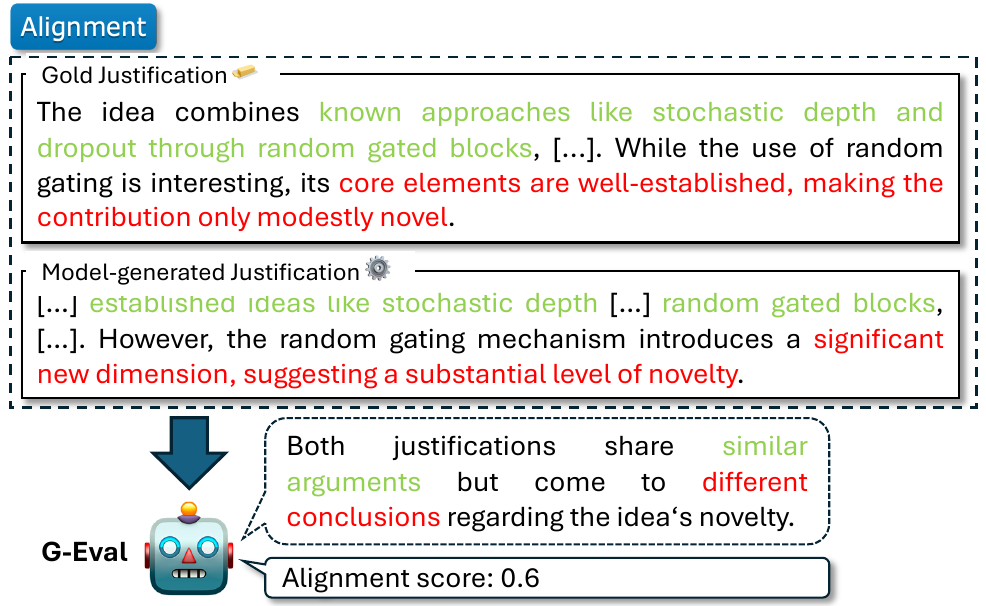}
    \caption{Evaluation of justification \textit{alignment} for novelty judgments using the G-Eval framework, which produces textual reasoning and a numerical score. We use only the numerical score for evaluation.}
    \label{fig:alignment_metric}
\end{figure}

\paragraph{Alignment} Evaluating the alignment of novelty judgment justifications is essential to ensure that a model's decision-making process mirrors human-like judgment, both in terms of logic and argumentation. This metric evaluates whether the reasoning behind a predicted novelty judgment is consistent with the reasoning in the gold standard. Specifically, it verifies whether a model-generated justification follows the same line of argumentation, presents similar supporting arguments, and reaches the same conclusion as the human gold justification. As illustrated in Figure \ref{fig:alignment_metric}, we utilize the G-Eval framework \cite{liu-etal-2023-g} to prompt an \ac{llm} for alignment evaluation, generating an alignment score that ranges from 0 (worst) to 1 (best).

\begin{figure}[h!]
    \centering
    \includegraphics[width=1\columnwidth]{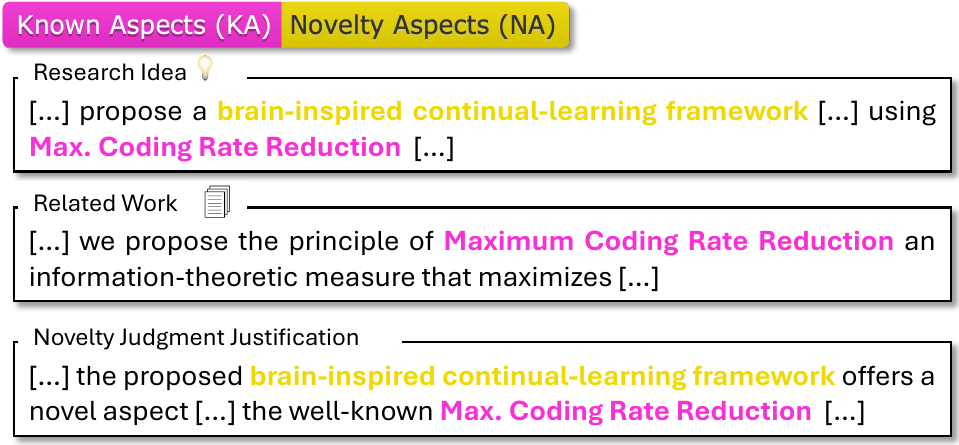}
    \caption{Example illustrating \textit{known} and \textit{novelty} aspects in novelty judgment justifications. \textcolor[HTML]{f731d6}{\textit{Known aspects}} refer to elements in a justification that highlight already established concepts or findings from previous work in a research idea. \textcolor[HTML]{EEDA00}{\textit{Novelty aspects}} denote elements in a justification that highlight new contributions of a research idea, which do not exist in prior work.}
    \label{fig:aspects_example}
\end{figure}

\begin{figure}[h!]
    \centering
    \includegraphics[width=1\columnwidth]{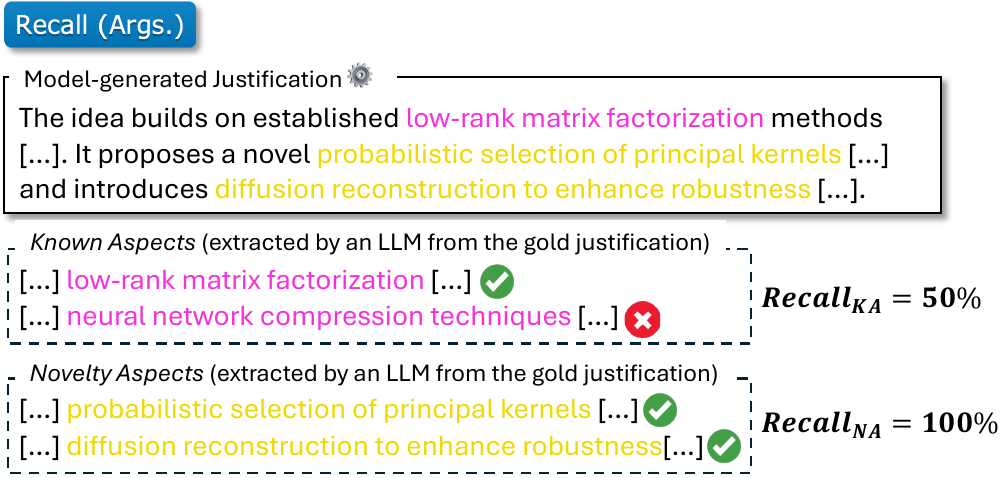}
    \caption{Example of \textit{Known Aspects Recall} and \textit{Novelty Aspects Recall} for evaluation of novelty judgment justifications.}
    \label{fig:recall_metric}
\end{figure}

\paragraph{Known Aspects Recall} We measure the extent to which arguments in the gold novelty justification that pertain to ``\textit{known aspects}'' (see Figure \ref{fig:aspects_example}) are captured in a model-generated justification. Following the FActScore approach \cite{min-etal-2023-factscore}, an \ac{llm} first extracts all arguments from both the model-generated and gold justifications. It then verifies whether each argument extracted from the model-generated justification is supported by the gold justification. The final metric is computed as:

\begin{equation}
\text{Recall}_{\text{Args.}} = 
\begin{cases} 
\min\left(1, \frac{N_{\text{supp}}}{N_{\text{gold}}}\right) & \text{if } N_{\text{gold}} > 0, \\
0 & \text{if } N_{\text{gold}} = 0.
\end{cases}
\label{eq:recall}
\end{equation}

where $N_{\text{supp}}$ is the number of model-generated known-aspect arguments supported by the gold justification, and $N_{\text{gold}}$ is the total number of known-aspect arguments in the gold justification. Figure \ref{fig:recall_metric} provides an illustrated example.

\paragraph{Novelty Aspects Recall} Analogous to above, we measure the extent to which arguments in the gold novelty justification related to ``\textit{novelty aspects}'' (see Figure \ref{fig:aspects_example}) are captured by a model-generated justification. This is done using the same \ac{llm}-based argument extraction and validation approach as in \textit{Known Aspects Recall}, with the metric computed using Equation \ref{eq:recall}. In this case, $N_{\text{supp}}$ represents the number of model-generated novelty-aspect arguments supported by the gold justification, while $N_{\text{gold}}$ denotes the total number of novelty-aspect arguments in the gold justification.

\begin{figure}[ht]
    \centering
    \includegraphics[width=1\columnwidth]{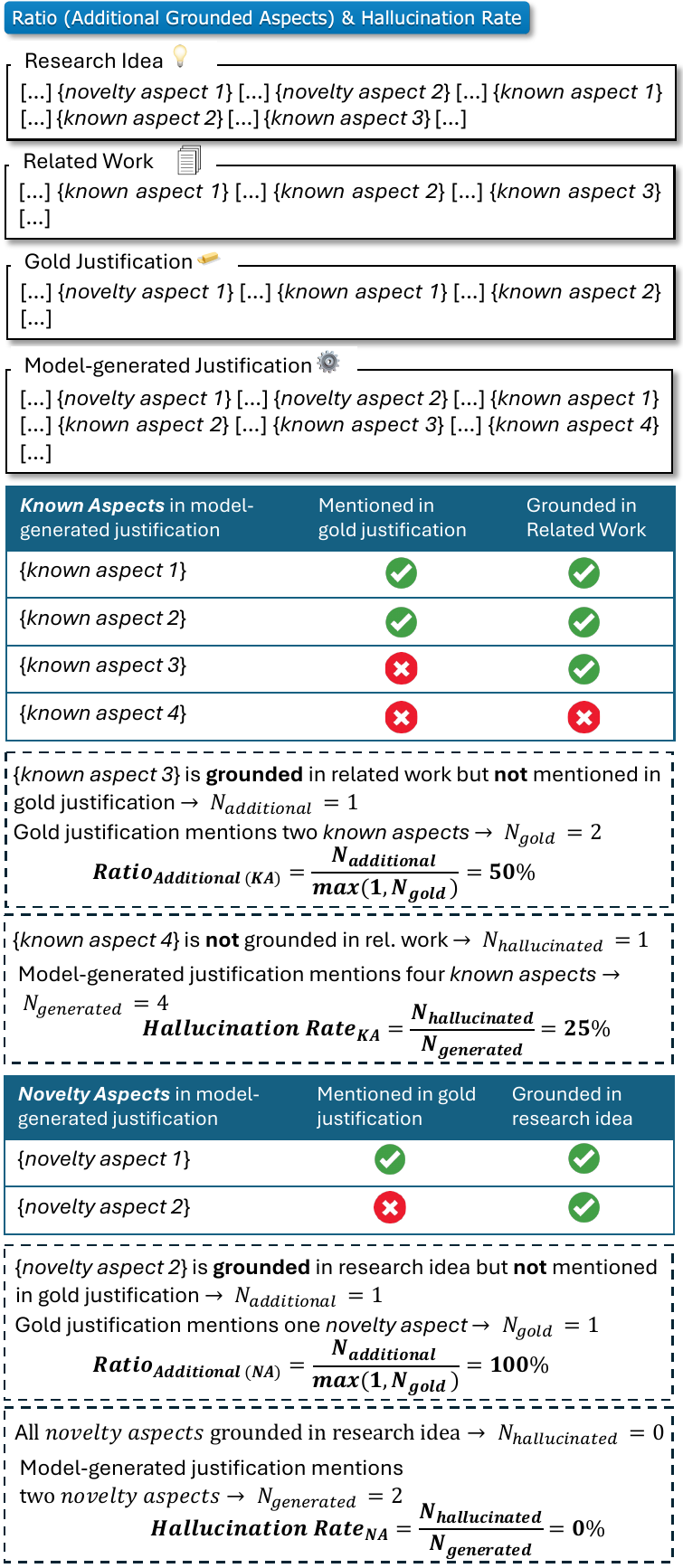}
    \caption{Example evaluation of a model-generated novelty judgment justification using \textit{Additional Ratio} and \textit{Hallucination Rate} for \textit{known aspects} and \textit{novelty aspects} respectively.}
    \label{fig:add_hall_metrics}
\end{figure}

\paragraph{Additional Known Aspects Ratio} We measure the extent to which a model generates additional known-aspect arguments, which are not present in the gold justification but grounded in the associated related works. To evaluate this, we again use the same \ac{llm}-based argument extraction and validation approach as in \textit{Known Aspects Recall}. Additionally, the \ac{llm} verifies whether the known-aspect arguments extracted from the model-generated justification are grounded in the corresponding related works. The metric is then computed as:

\begin{equation}
\text{Ratio}_{\text{Additional}} = \frac{N_{\text{additional}}}{\max(N_{\text{gold}},1)}
\label{eq:ratio}  
\end{equation}

where $N_{\text{additional}}$ is the number of model-generated known-aspect arguments unsupported by the gold justification but grounded in the related works and $N_{\text{gold}}$ is the total number of known-aspect arguments in the gold justification. Figure \ref{fig:add_hall_metrics} provides an illustrated example.

\paragraph{Additional Novelty Aspects Ratio} Similarly to above, we assess the extent to which a model generates additional novelty-aspect arguments that are not present in the gold justification but are grounded in the respective research idea. This is achieved using the same \ac{llm}-based argument extraction and validation approach as in \textit{Known Aspects Recall}, with an added \ac{llm}-based step to verify whether the extracted novelty-aspect arguments from the model-generated justification are grounded in the corresponding research idea. The metric is computed using Equation \ref{eq:ratio}, where $N_{\text{additional}}$ is the number of model-generated novelty-aspect arguments unsupported by the gold justification but grounded in the research idea and $N_{\text{gold}}$ is the total number of novelty-aspect arguments in the gold justification. 

\paragraph{Known Aspects Hallucination Rate} We quantify the extent to which model-generated justifications contain hallucinated known-aspect arguments (i.e., those not supported by any of the corresponding related works). To this end, we adopt the same \ac{llm}-based argument extraction and validation approach used in \textit{Known Aspects Recall} and compute the metric as:

\begin{equation}
\text{Hall. Rate} = 
\begin{cases} 
\frac{N_{\text{hallucinated}}}{N_{\text{generated}}} & \text{if } N_{\text{generated}} > 0, \\
0 & \text{if } N_{\text{generated}} = 0.
\end{cases}
\label{eq:hallucination_rate}
\end{equation}

where $N_{\text{hallucinated}}$ is the number of model-generated known-aspect arguments unsupported by any of the corresponding related works and $N_{\text{generated}}$ is the total number of known-aspect arguments in the model-generated justification. Figure \ref{fig:add_hall_metrics} provides an illustrated example.

\paragraph{Novelty Aspects Hallucination Rate} Similarly, we quantify the extent to which model-generated justifications contain hallucinated novelty-aspect arguments (i.e., those unsupported by the corresponding research idea). Using the same approach as in \textit{Known Aspects Hallucination Rate}, we compute this metric with Equation \ref{eq:hallucination_rate}, where $N_{\text{hallucinated}}$ represents the number of model-generated novelty-aspect arguments unsupported by the research idea and $N_{\text{generated}}$ is the total number of novelty-aspect arguments in the model-generated justification. 

For all \ac{llm}-based evaluations, it is crucial to use a high-performing model to ensure the accuracy of the various metrics. Accordingly, we use the GPT-4.1 \cite{OpenAI_2025} model for all \ac{llm}-based evaluations. Additionally, each presented justification metric is computed per individual sample. To provide a comprehensive evaluation in \acs{rino}, we average the computed justification metric scores across multiple samples.

\section{Benchmarking \acp{llm} as Judges of Research Idea Novelty}\label{sec:benchmarking}

\begin{table*}[ht!]
\centering
\renewcommand{\arraystretch}{1.2} 
\resizebox{\textwidth}{!}{
\begin{tabular}{|l|c|c|c|c|c|c|c|c|c|c|c|c|c|c|c|}
\hline
 & \multicolumn{7}{c|}{Novelty Score Metrics} & \multicolumn{7}{c|}{Justification Metrics} \\
\cline{2-15} 
\multirow{2}{*}{\textbf{Model}} & \multicolumn{6}{c|}{$\mathbf{F}_{\mathbf{1}}$} & \multirow{2}{*}{\textbf{MAE}} & \multirow{2}{*}{\textbf{ALI}} & \multicolumn{2}{c|}{\textbf{Recall}} & \multicolumn{2}{c|}{\textbf{Add. Ratio}} & \multicolumn{2}{c|}{\textbf{Hall. Rate}} \\
\cline{2-7} \cline{10-15}
 & Macro & 1 & 2 & 3 & 4 & 5 & & & KA & NA & KA & NA & KA & NA\\
\hline
\multicolumn{15}{|c|}{\textbf{Non-Reasoning Models}} \\
\hline
Llama-3.1-8B & 14.6 & 0.0 & 0.0 & 26.2 & 41.3 & \textbf{5.4} & 1.00 & 0.58 & 85.5 & 75.3 & 62.8 & 111.6 & 4.2 & 3.4 \\
Llama-3.3-70B & 9.5 & 0.0 & 0.0 & 2.2 & 45.0 & 0.0 & 1.04 & 0.55 & 88.9 & 78.3 & 86.3 & 115.3 & 1.1 & 1.4\\
Llama-4-Scout & 13.0 & 0.0 & 0.0 & 17.1 & 42.7 & 5.1 & 1.01 & 0.58 & 89.8 & 81.9 & 89.0 & \textbf{120.3} & \textbf{0.0} & 1.1\\
\hline
\multicolumn{15}{|c|}{\textbf{Reasoning Models}} \\
\hline
GPT-OSS-120B & 14.6 & 0.0 & 3.0 & 30.1 & 40.7 & 0.0 & 0.96 & 0.64 & 88.1 & 77.8 & 79.0 & 92.4 & 0.9 & 0.5 \\
DeepSeek-R1 & 12.3 & 0.0 & 0.0 & 16.1 & 45.6 & 0.0 & 0.99 & 0.67 & 87.8 & 81.1 & 115.7 & 112.4 & 0.6 & \textbf{0.2}\\
o3 & 16.2 & 0.0 & 5.6 & \textbf{35.6} & \textbf{39.7} & 0.0 & \textbf{0.93} & \textbf{0.72} & \textbf{90.4} & 85.6 & \textbf{139.9} & 74.0 & 1.3 & 1.7 \\
GPT-5 & \textbf{17.2} & 0.0 & \textbf{16.7} & 32.2 & 37.3 & 0.0 & \textbf{0.93} & 0.71 & 89.9 & \textbf{85.7} & 122.1 & 91.8 & 0.6 & 0.5 \\
\hline
\end{tabular}
}
\caption{Evaluation results of novelty judgments on the \acs{rino} test set. As described in Section \ref{sec:novelty_score_metrics}, the reported novelty score metrics include \textit{$F_{1}$} macro averaged and for each rubric category (1-5), as well as \textit{\ac{mae}}. Further, as outlined in Section \ref{sec:justification_metrics}, the justification metrics include \textit{Alignment (ALI)}, as well as \textit{Recall}, \textit{Additional Ratio} (in \%), and \textit{Hallucination Rate} (in \%) for \textit{Known Aspects (KA)} and  \textit{Novelty Aspects (NA)} respectively.}
\label{tab:experiment_results}
\end{table*}


In this section, we present a benchmarking study examining several state-of-the-art \acp{llm} on their ability to judge the novelty of research ideas. To this end, we follow recent works that frame the novelty judgment of research ideas as zero-shot task by directly giving the review criteria and prompting \acp{llm} for a final score \cite{yang-etal-2024-large-language,lu2024aiscientistfullyautomated,si2025can,li2025mlrcopilotautonomousmachinelearning,baek-etal-2025-researchagent}. Accordingly, we instruct each \ac{llm} to perform the \acs{rino} task illustrated in Figure \ref{fig:task_example} to generate a numerical novelty score and a textual justification. Thereby, the \ac{llm} is provided with the novelty judgment rubric detailed in Table \ref{tab:novelty-rubric}, alongside a research idea and its related works. The \ac{llm} is then asked to analyze the research idea, identify its key contributions, and compare it to the provided related works. Based on this analysis and comparison, the model is finally tasked to generate a suitable novelty score according to the rubric, accompanied by a brief justification explaining its reasoning for the predicted  score. The exact instructions are shown in Figure \ref{fig:zs-prompt}. 

For this study, we select a diverse set of \acp{llm}, encompassing a range of sizes and reasoning capabilities. Specifically, we include non-reasoning models including \textit{Llama-3.1-8B} \cite{grattafiori2024llama3herdmodels}, \textit{Llama-3.3-70B} \cite{grattafiori2024llama3herdmodels}, and \textit{Llama-4-Scout-17B-16E} \cite{Meta_2025}, as well as reasoning-capable models including \textit{DeepSeek-R1} \cite{deepseekai2025deepseekr1incentivizingreasoningcapability}, \textit{GPT-OSS-120B} \cite{openai2025gptoss120bgptoss20bmodel}, \textit{o3} \cite{OpenAI_2025b}, and \textit{GPT-5} \cite{OpenAI-gpt-5}. This selection enables a comprehensive evaluation of model performance across different architectures and reasoning abilities.

\paragraph{Novelty Judgment Performance} Table~\ref{tab:experiment_results} shows the evaluation results. We observe that all models show very low novelty judgment abilities, with none of them achieving significant $F_{1}$ scores and the highest macro-average being 17.2. Notably, no model successfully predicted the novelty category 1, as indicated by the 0.0 $F_{1}$ scores for this category across all models. This suggests a strong bias against predicting ideas as "\textit{not novel}", indicating that the models tend to avoid judging ideas as lacking novelty altogether. Interestingly, novelty categories 2 and 5 are occasionally predicted correctly, but the models predominantly predicted novelty categories 3 and 4. This suggests that the models tend to avoid assigning extreme values of novelty (such as "\textit{marginally novel}" or "\textit{highly novel and innovative}"). Instead, they consistently attempt to find at least some aspect of novelty in a research idea, even if it is not present. The \ac{mae} values are relatively low and consistently hover around 1, indicating that while the models may often make errors in their novelty judgments, these do not deviate drastically from the gold standard.

\paragraph{Quality and Coverage of Justifications} The justification metrics provide interesting insights into how the models substantiate their novelty judgments. It is noteworthy that all models exhibit relatively high recall, implying that the \acp{llm} frequently use arguments similar to those found in the human-annotated gold standard justifications. This finding is consistent with the results of \citet{afzal2026notnovelenoughenriching}, who reported high alignment between \ac{llm} and human novelty reasoning. Moreover, the high additional ratios (Add. Ratio), often exceeding 100\%, suggest that the \acp{llm} tend to generate more elaborate justifications than humans and frequently find more arguments to justify their novelty predictions. This may be because, for humans, one or two well-chosen arguments are often sufficient to judge the novelty of a research idea, while \acp{llm} appear to strive to provide a comprehensive set of arguments, likely in an attempt to satisfy the user. 

\paragraph{Hallucinations and Judgment–Justification Gap} Despite generating many arguments, the hallucination rate is low across all models, suggesting that the models' justifications are largely grounded in the provided context. This indicates that, while the models may sometimes over-elaborate in their novelty judgment justifications, the arguments they generate are mostly reliable and supported by evidence. This stands in contrast to their novelty score predictions, which are more dissimilar from the human-annotated gold novelty scores, pointing to a gap in the models' ability to accurately judge the novelty of research ideas, even if their justifications seem to contain plausible arguments.

\paragraph{Reasoning vs. Non-Reasoning Models} When comparing model performance, we observe that reasoning models outperform their non-reasoning counterparts, albeit by a small margin. The GPT-5 model achieves the highest macro-averaged $F_{1}$ score with 17.2, closely followed by o3 (16.2). These models, designed for more complex reasoning tasks, outperform the non-reasoning models, which generally exhibit lower $F_{1}$ scores and demonstrate worse novelty judgment abilities. This indicates that incorporating additional reasoning and deeper thinking during the generation process helps models make more accurate judgments regarding the novelty of research ideas.

\paragraph{Takeaway} Overall, the results show that the \ac{llm}-generated novelty judgment justifications closely align with those of human experts, whereas the predicted novelty judgment scores diverge substantially from the human-annotated gold scores, pointing to a clear gap: \textit{although \acp{llm} can generate plausible and well-supported novelty justifications, they fail to translate their reasoning into accurate novelty judgments}. Further, the  models tend to avoid extreme predictions and avoid judging ideas as \textit{not novel at all} nor \textit{highly novel and innovative}. Instead, they constantly strive to find a middle ground. Additionally, while the models' justifications are often grounded in the provided context, they tend to be more elaborate than human justifications, reflecting a difference in how humans and models approach novelty judgment. Despite these differences, the models' predictions are only slightly different from human annotations (see \ac{mae} scores), suggesting that their beliefs about novelty do not differ fundamentally, but are instead somewhat inaccurate or imprecise. 

\section{Conclusion}
This work introduces \acs{rino}, the first automated benchmark for evaluating novelty judgments of research ideas. It includes 1,381 research ideas derived from and judged by human experts. Further, the benchmark comprises nine automatic metrics to assess the accuracy of predicted novelty scores and to compare model-generated textual justifications with human-authored gold justifications. Our work bridges the gap between current, largely manual and incomparable human evaluations, towards reproducible and comparable evaluations.

Further, we investigate the capability of several state-of-the-art \acp{llm} to judge the novelty of research ideas. Our findings reveal that current \acp{llm} face substantial challenges in accurately judging the novelty of research ideas. Notably, while \acp{llm} refrain from judging ideas as completely lacking novelty, they tend to seek a middle ground, aiming to attribute at least some degree of novelty while simultaneously avoiding judgments of high novelty and innovation. Interestingly, as indicated by the strong correspondence between \ac{llm}-generated and human-authored justifications, \ac{llm} reasonings align closely with human rationales for research idea judgment. However, this alignment does not translate to the \ac{llm}-predicted novelty scores, which diverge considerably from human-assigned scores. Finally, while all \acp{llm} examined exhibit difficulties in effective novelty judgment, our experiments indicate that reasoning-capable models consistently outperform non-reasoning ones, suggesting that longer thinking and deeper engagement with the input and task instructions improve \ac{llm}-based novelty judgment of research ideas.



\section{Limitations}

\acs{rino} is derived exclusively from ICLR 2022 and 2023 submissions, limiting its domain, epistemic, and cultural diversity. Because it reflects a single conference ecosystem centered on machine learning, the benchmark captures the reviewing norms and novelty criteria of that community. Fields with different epistemological assumptions and evaluation practices—such as many areas in the humanities and social sciences—are not represented. Consequently, the novelty dimensions emphasized in \acs{rino}, focused on technical or methodological innovation, may underrepresent theoretical or discovery-oriented contributions. Findings based on \acs{rino} should therefore be interpreted within this specific context and validated on broader, more heterogeneous datasets.

The dataset relies on peer review scores, which are inherently subjective and shaped by disciplinary conventions. While this enables modeling real-world novelty judgments, it also operationalizes a particular reviewing culture rather than a universal notion of novelty, potentially reflecting individual or systemic biases.

Because the data originates from predominantly English-language OpenReview submissions, linguistic and rhetorical conventions may influence how novelty is expressed and assessed. Furthermore, although \acp{llm} are used to extract structured ideas and synthesize justifications with careful filtering, they may introduce hallucinations, normalization effects, or discourse-sensitive distortions.

Finally, \acs{rino} focuses solely on novelty and does not capture other dimensions of research quality, such as rigor, significance, or reproducibility.

\section{Ethics Statement}
All data in \acs{rino} is derived from publicly available OpenReview submissions and papers indexed by Semantic Scholar. No private reviewer or author information is included. We emphasize that the dataset is intended for research and educational purposes only. Users should not use models trained on this data to make formal or high-stakes judgments of research ideas, as novelty judgments are inherently subjective and context-dependent.



\section{Broader Impact}
\acs{rino} is designed to advance AI-assisted scientific discovery by enabling models to reason about and explain novel contributions in research. It provides a valuable resource for researchers, educators, and students to understand and teach research idea novelty judgment and serves as a benchmark for developing models with explainable reasoning capabilities.

At the same time, automated predictions of research idea novelty should not replace human expert judgment. The dataset reflects inherently subjective human opinions and may contain biases from peer reviews. Models developed on \acs{rino} should be intended as tools to support, rather than replace human judgments of research ideas.

\section{Acknowledgements}
The authors acknowledge the financial support by the Federal Ministry of Research, Technology and Space of Germany and by Sächsische Staatsministerium für Wissenschaft, Kultur und Tourismus in the programme Center of Excellence for AI-research „Center for Scalable Data Analytics and Artificial Intelligence Dresden/Leipzig“, project identification number: ScaDS.AI.

The first author was supported by a scholarship of the German Academic Exchange Service (DAAD). 

We used AI-based assistance tools to support language editing, minor formatting, and coding tasks. These tools did not contribute to the intellectual content or scientific conclusions. All content was reviewed by the authors, who assume full responsibility for the publication.

\section{Bibliographical References}\label{sec:reference}

\bibliographystyle{lrec2026-natbib}
\bibliography{custom-bibliographic}


\appendix
\section{Appendix}\label{sec:appendix}

\begin{figure}[t!]
\centering
\footnotesize
\begin{mycode}[]{Research Idea Novelty Judgment Prompt}
You are an expert in machine learning research evaluation. You will be given two inputs:

1. A research idea with objective, problem statement, and solution approach.
2. A list of related works, each with a title and abstract.

Your task is to **assess the novelty of the research idea** compared to the related works.

### Instructions:
- Analyze the research idea and summarize its key contributions.
- Compare it with the related works to identify overlaps and differences.
- Specifically, assess whether the idea introduces **significant new aspects** not present in existing work, or if it is largely a variation on known approaches.
- Provide your output as a **JSON object only**, with:
  - "reasoning": a short paragraph (2-4 sentences) explaining the reasoning behind the novelty score.
  - "novelty_score": an integer between 1-5 where: {novelty_rubric}
  
### Inputs:

**Research Idea:**
{research_idea}

**Related Works:**
{related_works}

### Output Format:
```json
{{
  "reasoning": <short explanation>,
  "novelty_score": <1|2|3|4|5>
}}
\end{mycode}
\caption{Zero-shot instructions for judging the novelty of research ideas.}
\label{fig:zs-prompt}
\end{figure}

\end{document}